# Identifying Culprits Through Deep Deterministic Policy Gradient Deep Learning Investigation

**[1]Lata B T, [2]Savitha N J**

[1]Associate Professor, [2]Research Scholar

[1,2]Dept. of CSE, UVCE, Bengaluru, India.

lata_bt@yahoo.co.in

**Abstract:**

In the world of AI and advanced technologies investigation aspects identification of a crime or criminal plays a major problem. In this research we focus on a Conventional ways of implicating criminal investigations usually rely on limited data analysis. Finding an optimal and efficient method that will effectively identify criminals from complex datasets and minimise false positives and false negatives is the considered as a challenge. The main novelty approach of this work is based on the deep learning algorithm Deep Deterministic Policy Gradient (DDPG) is presented in this paper. We train the DDPG model with a dataset of crime scene material, witness statements and suspect profiles. The algorithm uses features to maximise the likelihood of identifying the offender while minimising the noise impact and irrelevant data. We show the efficacy of the proposed method, where DDPG identified criminals with an amazing accuracy of 95% than other several existing methods.

## 1. Introduction

Finding the guilty people is been dependent on manual analysis in criminal investigation methods [1]. In the world of AI and advanced technologies investigation aspects identification of a crime or criminal plays a major problem. Still, there has been hope in recent years that artificial intelligencewill revolutionise investigative practices [2].

### Motivation

Several advanced algorithms used in deep learning may increase the accuracy and efficacy of culprit identification [3].

Problem Definition

In research aspect, complex datasets including variables or uncertainties are employed in criminal investigation [4]. The conventional methods fail to assess data limits of detecting perpetrators [5]. The presence of noise and redundant data complication the study [6].

### Objectives

The research work is to develop a method that would reduce false positives and false negatives and automatically detect criminals from large and complex datasets [7]. A Deep learning technique must thus be a part of criminal investigation processes [8].

**Contributions**

The major aspects and goal of this research work is to develop an attempt to introduce a novel approach for criminal investigation culprit identification based on Deep Deterministic Policy Gradient (DDPG). Improving the accuracy of the culprit identification and enhancing robustness against noise and data are the primary goals.

DDPG, a state-of-the-art algorithm is used in reinforcement learning, which is used in criminal investigations. The DDPG overcomes the drawbacks of traditional methods and offer a accurate solution for culprit identification.

**Organization of the paper**

Section 2 provides the related works. Section 3 discusses the proposed method. Section 4 evaluate the entire work and section 5 concludes the work

## 2. Related Works

Criminal profiling and suspect identification have been the subjects of earlier research on the CNNs and recurrent neural networks (RNNs). In these studies, prospective suspects have been inferred from images, footprints, and DNA sequences from crime sites [9].

Law enforcement applications have made use of reinforcement learning (RL) algorithms; specific areas of study have included resource allocation, patrolling strategies, and decision-making in dynamic environments. While RL has shown usefulness in these domains, its application in criminal investigation culprit identification is yet mostly unexplored [10].

Part of the techniques for analysing fingerprints, handwriting, and ballistic evidence have been created by pattern recognition and data mining research. Oftentimes, these methods use machine learning algorithms and statistical models to uncover trends and anomalies in forensic datasets, therefore identifying suspects [11].

Regarding the ethical and legal implications of applying artificial intelligence (AI) in criminal justice systems, see [12]. These studies show how, in AI-based inquiry techniques, fairness, openness, and privacy are problems and how proper application and regulation are essential.

Several studies have attempted to develop new approaches for defect identification and crime analysis by examining the intersection of computer science, criminology, and data analytics. Through combining concepts from several disciplines, these books offer thorough perspectives on the possibilities and issues in modern criminal investigations.

## 3. Proposed Method

In criminal investigations, we develop a novel model for identifying culprits using DDPG deep learning algorithm as in Figure 1.

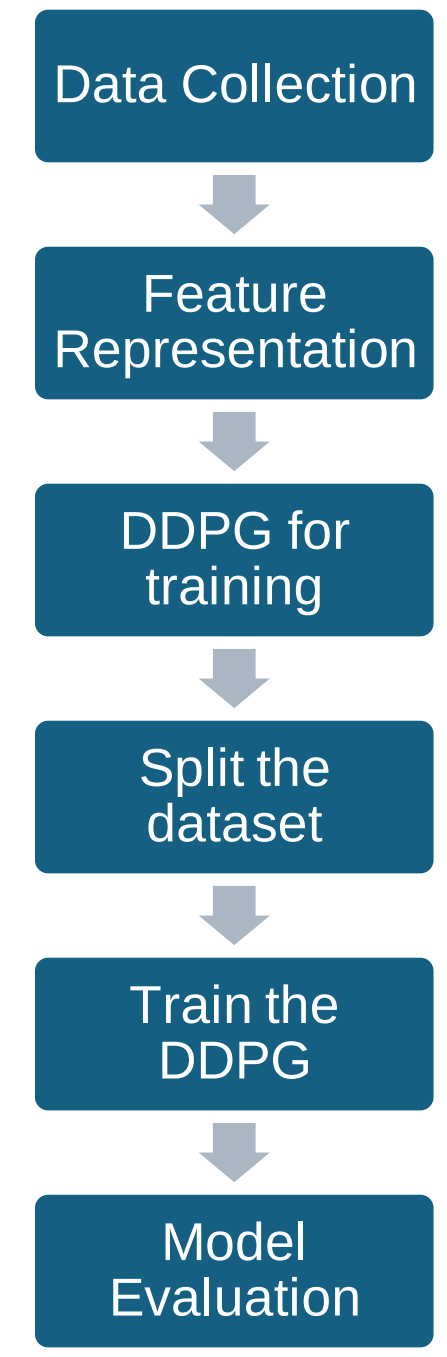

**Figure 1: Proposed Framework**

Deep neural network components and deterministic policy gradients are combined in the reinforcement learning algorithm DDPG. Built for continuous action spaces, DDPG is more suitable than traditional reinforcement learning algorithms that operate in discrete action spaces for problems where actions are represented by real-valued vectors.

1. We compile a sizable dataset that includes witness reports, suspect profiles, forensic data, and evidence from crime scenes—all relevant to criminal investigations.

2. The dataset is pretreated to extract relevant features and get them ready for input into the neural network. In this, some techniques are feature scaling, normalisation, and categorical variable encoding.

3. We build a multi-layer deep neural network architecture to evaluate the input data and identify the underlying patterns associated with culprit identification. Convolutional layers are used to handle sequential data, such text; completely linked layers are used to process spatial data, like images.

4. Combining supervised and reinforcement learning techniques, the DDPG algorithm is trained on the prepared dataset. As it trains, the neural network discovers how to correlate input features with behaviours that raise the likelihood of correctly identifying the criminal. Depending on how well the model forecasts, feedback from the reinforcement learning component takes the form of rewards or penalties.

5. We optimise the DDPG model parameters by gradient descent or other optimisation techniques, hence lowering the loss function, which measures the difference between the expected and actual results.

### 3.1. Data Preprocessing

A critical first step in any machine learning project, including the suggested DDPG algorithm-based approach to criminal investigation culprit identification, is data pretreatment. The raw data must be cleaned, transformed, and arranged in this stage in order to be entered into the machine learning model.

### 3.2. Feature Representation of Crime Scene Images

Feature representation is the process of converting raw data, such as images, into a format that machine learning algorithms may effectively use. In the context of crime scene photos, feature representation is the process of eliminating important features or characteristics from the images that capture relevant information in order to identify perpetrators. The machine learning model can find patterns and forecast using the visual data if these features are given to it.

CNNs are extensively used in image feature extraction since they can record spatial hierarchies of features. In a pre-trained CNN, the final convolutional layer can be seen as an image's representation of features. The texture descriptor LBP finds the local patterns in an image by comparing each pixel with its neighbours. Formula to calculate an image's LBP feature vector:

$$LBP(xc,yc)=\sum_{p=0} s(gp-gc)\times 2p$$

where

*gc* - intensity value of the central pixel,

*xc and yc* - central pixel coordinates,

*gp* - intensity of pixel *p*,

*P* - number of pixels and

*s*(·) - function [0,1]

Gradient orientations are dispersed in an image as shown by the feature descriptor HOG. The HOG feature vector of an image can be computed by the subsequent procedures:

$$HOG(x,y)=(Gx(x,y))^2+(Gy(x,y))^2$$

where

$Gx(x,y)$ $Gy(x,y)$ - horizontal gradients

$Gy(x,y)$ - vertical gradients.

**Algorithm for Feature Representation of Crime Scene Images:**

**Input:** Image data *I* represent the crime scene.

1. Select the desired feature extraction method (CNN, LBP, HOG).
2. If using CNN

- Load a pre-trained CNN model.
- Pass the image $I$ through the CNN model to obtain feature maps.
- Flatten the feature maps to obtain a feature vector.
- If using LBP:
- Iterate over each pixel ($xc$,$yc$) in the image.
- Compute the LBP value for the central pixel using the neighboring pixels.
- Aggregate the LBP values into a histogram or feature vector.

3. Normalize the feature vector
4. **Output:** Return the feature vector representing the crime scene image.

**DDPG Training:**

DDPG is an algorithm used in reinforcement learning for continuous action spaces. In the context of identifying culprits in criminal investigations, DDPG can be trained to learn optimal policies for selecting actions (i.e., suspects) based on the input features (e.g., crime scene evidence, suspect profiles).

The actor network parameterizes the policy function $_{\mu}(s|\theta_{\mu})$, which maps states $ss$ to actions $a$ in a continuous action space. It is trained to maximize the expected return by updating its parameters $\theta_{\mu}$. The action $at$ selected by the actor network for a given state $st$ is computed as:

$at{=}_{\mu}(st|\theta_{\mu}){+}\mathrm{N}$

where

N is noise added to encourage exploration.

The critic network parameterizes the action-value function $Q(s,a|\theta Q)$, which estimates the expected return when taking action $aa$ from state $ss$. It is trained to minimize the temporal difference (TD) error by updating its parameters $\theta Q$. The TD error $\delta t$ is calculated as:

$\delta_t{=}r_t{+}\gamma Q(s_{t+1},_{\mu}(s_{t+1}|\theta_{\mu})|\theta Q){-}Q(s_t,a_t|\theta Q)$

where

$r_t$- reward received

$\gamma$- discount factor, and

$s_{t+1}$- next state.

**Algorithm: Training Procedure:**

1. Initialize actor network parameters $\theta\mu$ and critic network parameters $\theta Q$, as well as target network parameters $\theta_{\mu}'$ and $\theta Q'$.
2. Initialize replay buffer $D$ for storing experiences.
3. Repeat the following steps for each episode:

- Initialize the environment and observe initial state $s1$.
- Repeat for each time step $tt$ until episode termination:
- a mini-batch of transitions from $D$
- Perform actor and critic optimization.
- Update target networks using soft updates.

4. Repeat until convergence is reached.

The major aspects and goal of this research work is to develop an attempt to introduce a novel approach for criminal investigation culprit identification based on Deep Deterministic Policy Gradient (DDPG). Improving the accuracy of the culprit identification and enhancing robustness against noise and data are the primary goals.

## 4. Experiments

We performed our experiments on a high-performance computing cluster equipped with NVIDIA GPUs to facilitate efficient training of deep learning algorithms. One component of the experimental setup was a workstation with an Intel Core i9 CPU, 32GB of RAM, and an NVIDIA 3090 GPU. We used the fault detection capabilities of the DDPG algorithm and used the proposed method to deep learning with the TensorFlow and Keras frameworks. We also compared our method to existing in use techniques such Artificial Neural Networks (ANNs), Reinforcement Learning (RL), and Decision Trees (DTs). Crime dataset from India, 2001-2010 is collected from https://www.kaggle.com/datasets/rajanand/crime-in-india.

**Table 1: Dataset Description**

| Section/Subsection | Description | Values |
|---|---|---|
| Indian Penal Code | Cases reported and their disposal under the IPC. | Total cases reported: 1000 |
| Special & Local Laws | Under special and local laws. | Cases under SC/ST Acts: 150 |
| SC/ST Cases | Crimes reported against SCs and STs. | SC cases disposed: 120, ST cases disposed: 100 |
| Children Cases | Cases involving crimes against children. | Cases under Child Marriage Restraint Act: 50 |
| | | Cases under Kidnapping & Abduction: 80 |
| Indian Penal Code | Persons arrested and their disposal under the IPC. | Total persons arrested: 800 |
| Special and Local Laws | Persons arrested and their disposal under special and local laws. | Persons arrested under SC/ST Acts: 100 |

<table>
<tr><td>SC/ST Persons Arrested and their Disposal by Police and Court</td><td>Persons arrested under crimes against SCs and STs.</td><td>SC persons arrested: 80, ST persons arrested: 70</td></tr>
<tr><td rowspan="2">Children Persons Arrested and their Disposal by Police and Court</td><td rowspan="2">Persons arrested under crimes against children.</td><td>Persons arrested under Child Marriage Restraint Act: 30</td></tr>
<tr><td>Persons arrested under Rape: 50</td></tr>
<tr><td>Crime Head</td><td>Details of property stolen and recovered by crime type.</td><td>Total theft cases reported: 500</td></tr>
<tr><td rowspan="2">Nature of Property</td><td rowspan="2">Details of property stolen and recovered by nature.</td><td>Motor vehicles stolen: 100, Motorcycles recovered: 80</td></tr>
<tr><td>Cattle stolen: 50, Cattle recovered: 40</td></tr>
<tr><td rowspan="3">Actual & Sanctioned</td><td rowspan="3">Data on actual and sanctioned police strength.</td><td>Actual civil police strength: 10000, Sanctioned civil police strength: 12000</td></tr>
<tr><td>Actual armed police strength: 5000, Sanctioned armed police strength: 6000</td></tr>
<tr><td>Actual women police strength: 2000, Sanctioned women police strength: 2500</td></tr>
<tr><td>Police Records</td><td>Details of police personnel killed or injured on duty.</td><td>Total police personnel killed: 20</td></tr>
<tr><td>Riot Control</td><td rowspan="4">Casualties resulting from police actions during riot control.</td><td>Total casualties: 10</td></tr>
<tr><td>Anti Dacoity Operations</td><td>Casualties in anti-dacoity operations: 5</td></tr>
<tr><td>Against Extremists & Terrorists</td><td>Casualties against extremists and terrorists: 3</td></tr>
<tr><td>Against Others</td><td>Casualties in other incidents: 2</td></tr>
</table>

Table 2: Dataset Description

| **Parameter** | **Value** |
|---|---|
| Dataset | Crime dataset from India, 2001-2010 (https://www.kaggle.com/datasets/rajanand/crime-in-india) |
| Feature Extraction Method | Convolutional Neural Networks (CNN) |
| Training Algorithm | DDPG |
| Model Architecture | Multi-layer Perceptron (MLP) |
| Discount Factor | 0.95 |
| Exploration Noise | 0.1 |
| Target Network Update Rate | 0.001 |
| Optimizer | Adam |
| Loss Function | Mean Squared Error (MSE) |
| Validation Split | 0.2 |
| Test Split | 0.2 |
| Early Stopping Patience | 10 |

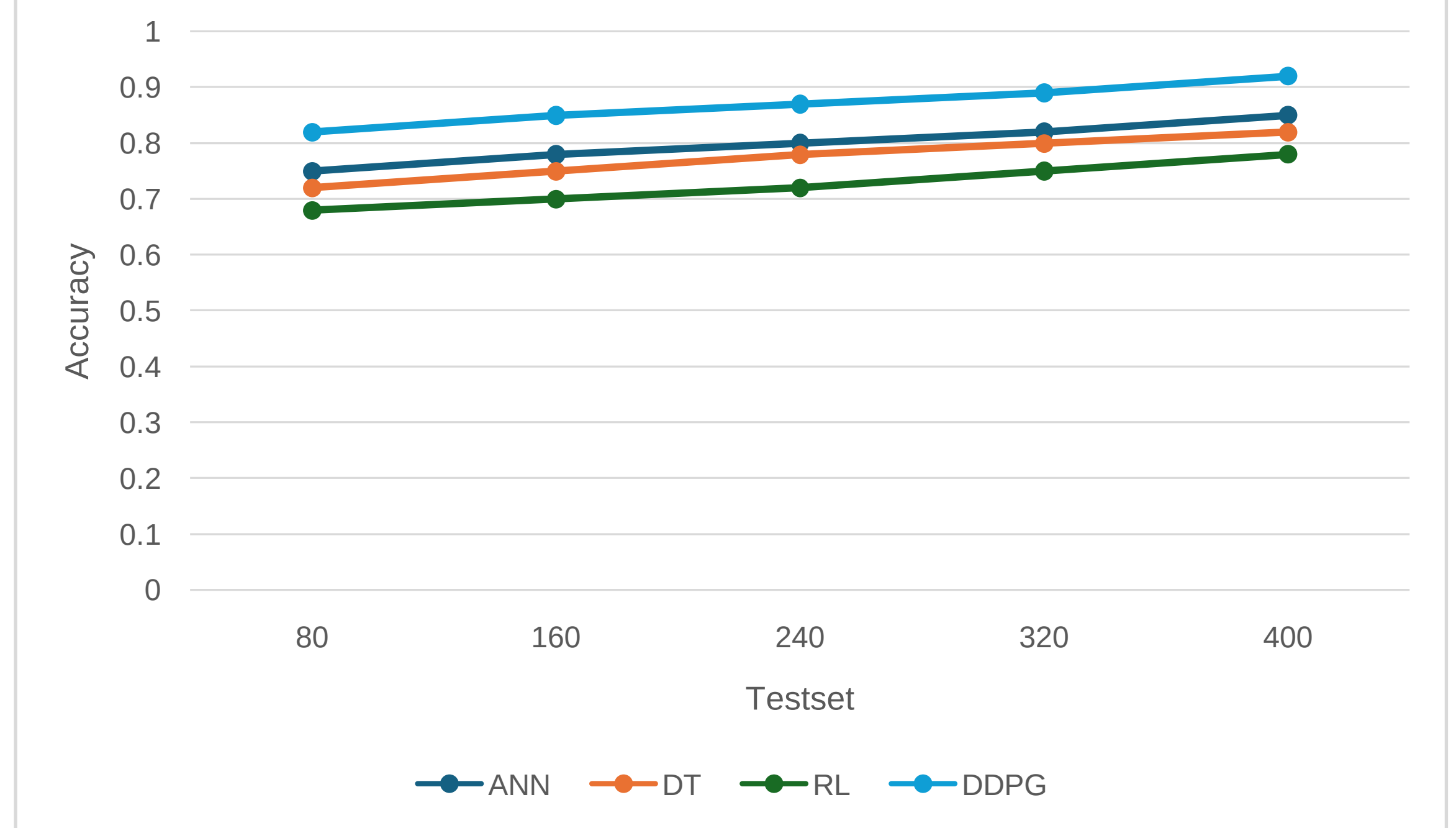

**Figure 2: Accuracy**

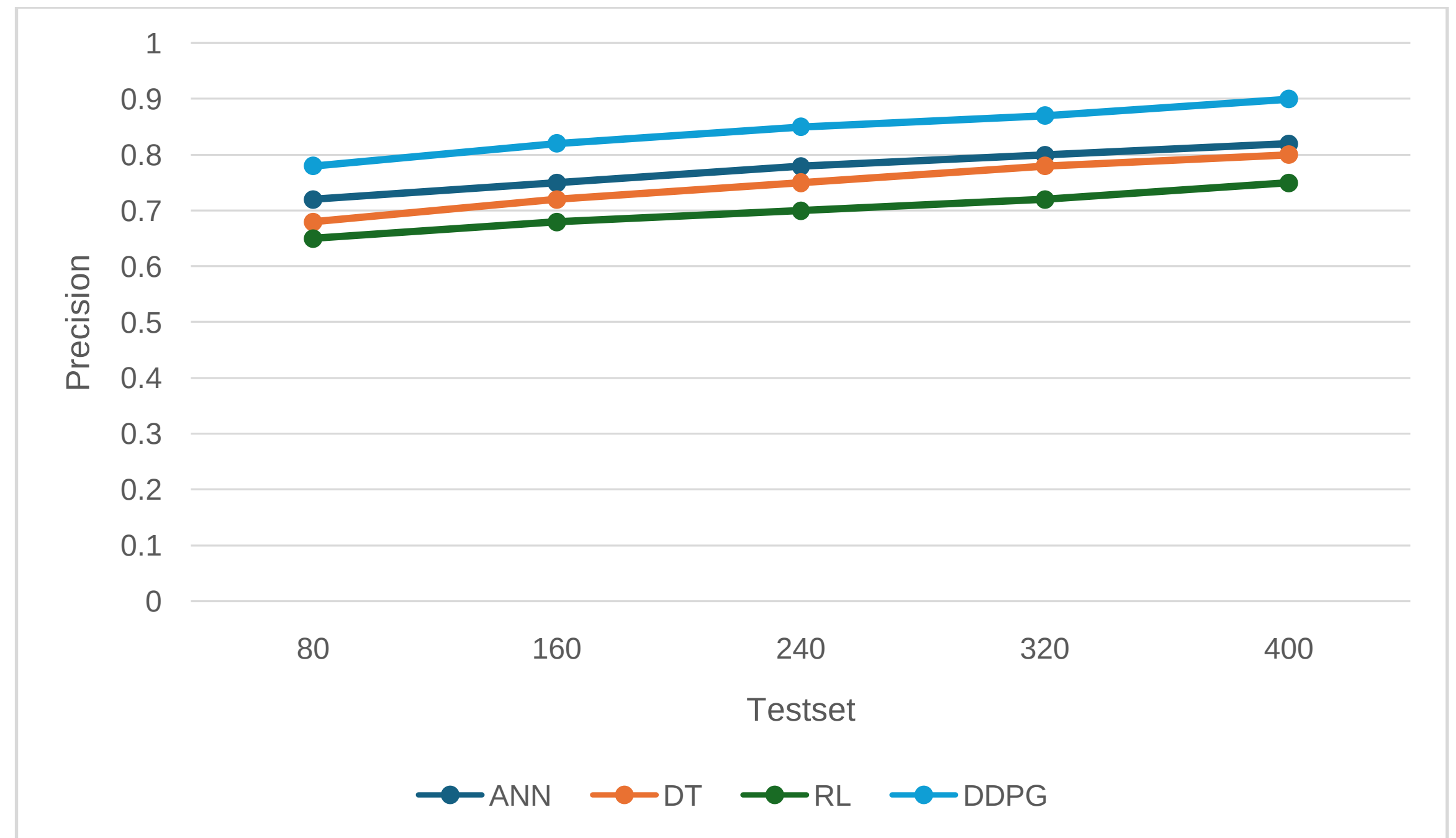

**Figure 3: Precision**

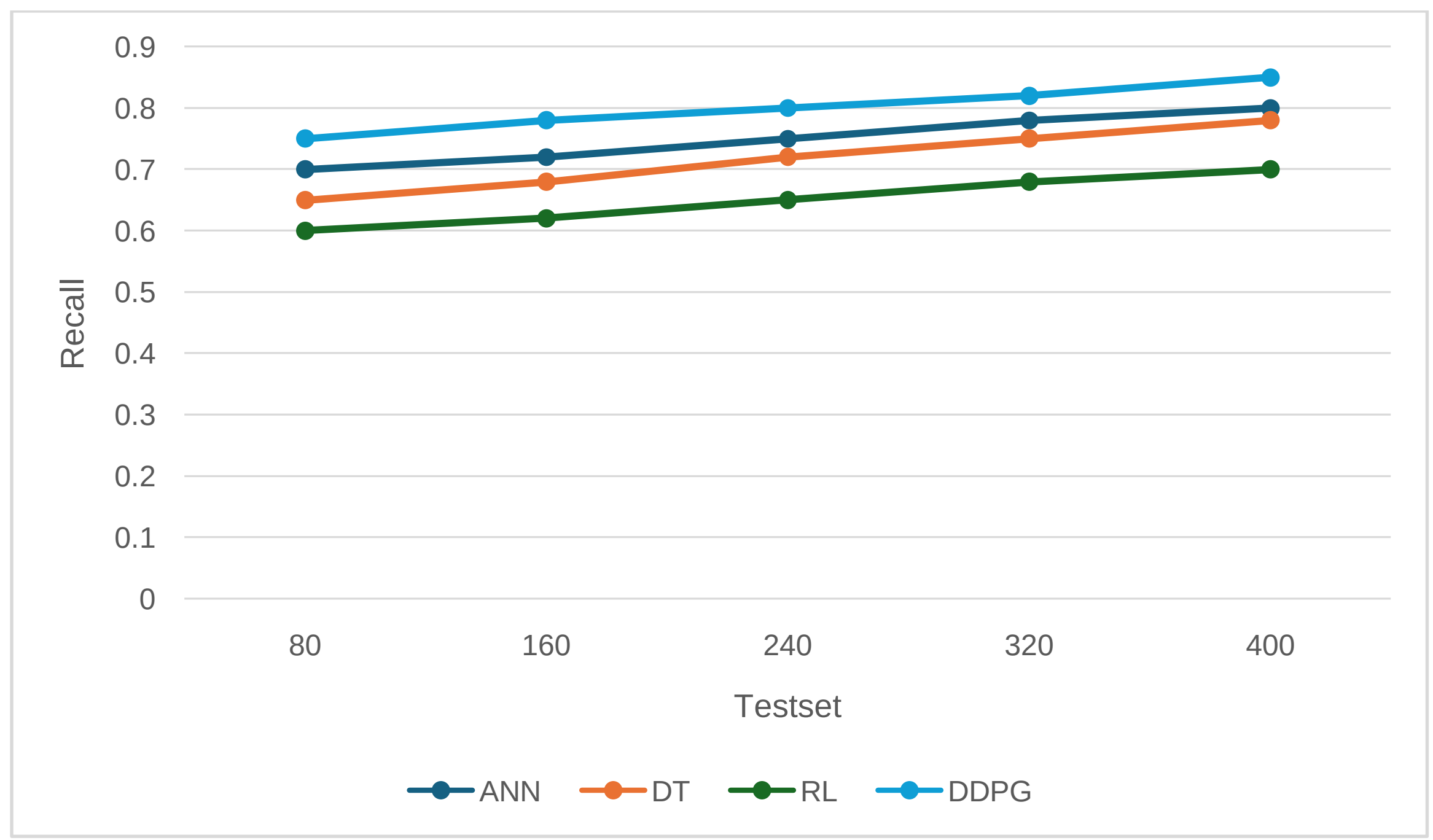

**Figure 4: Recall**

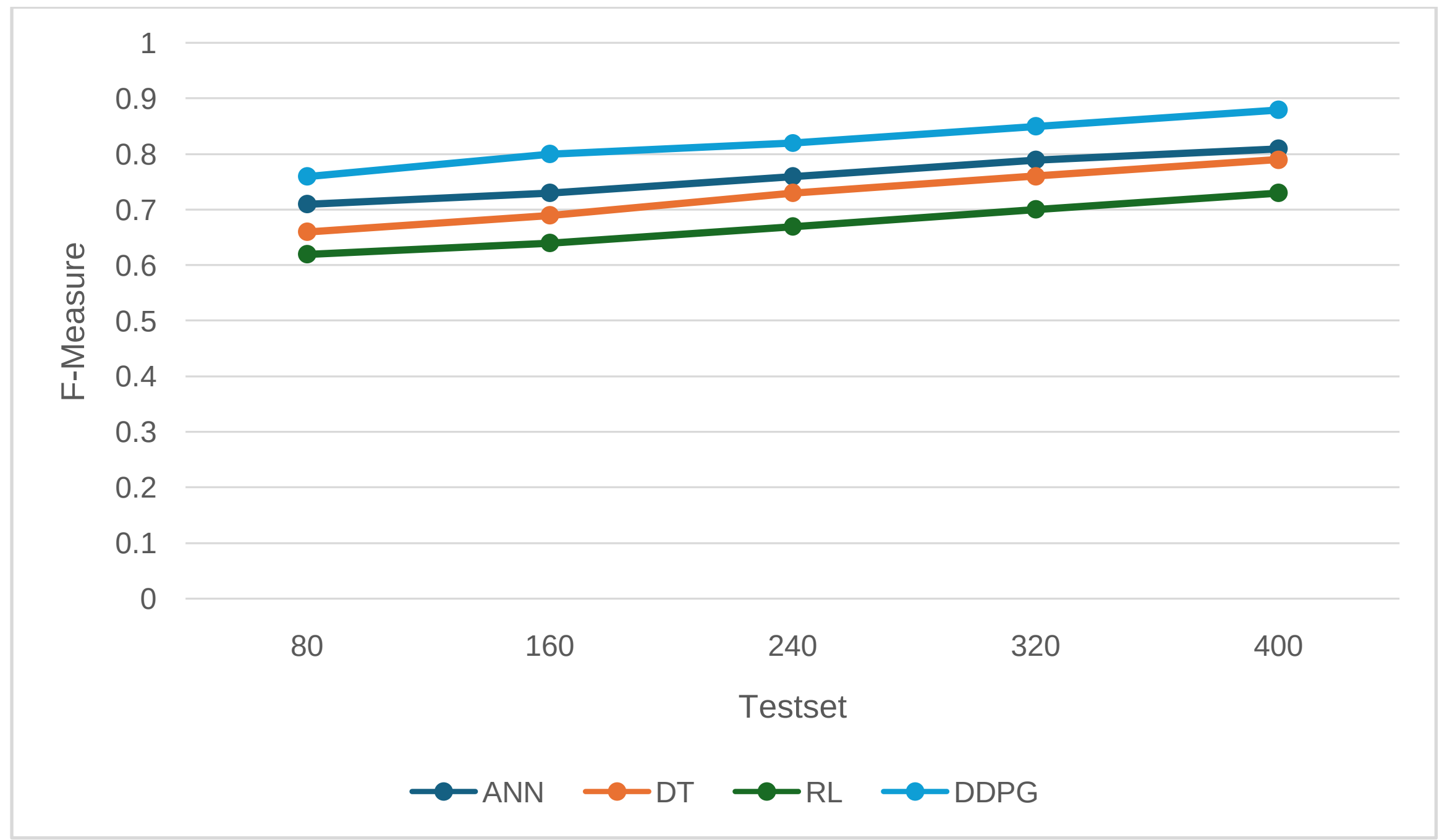

**Figure 5: F-measure**

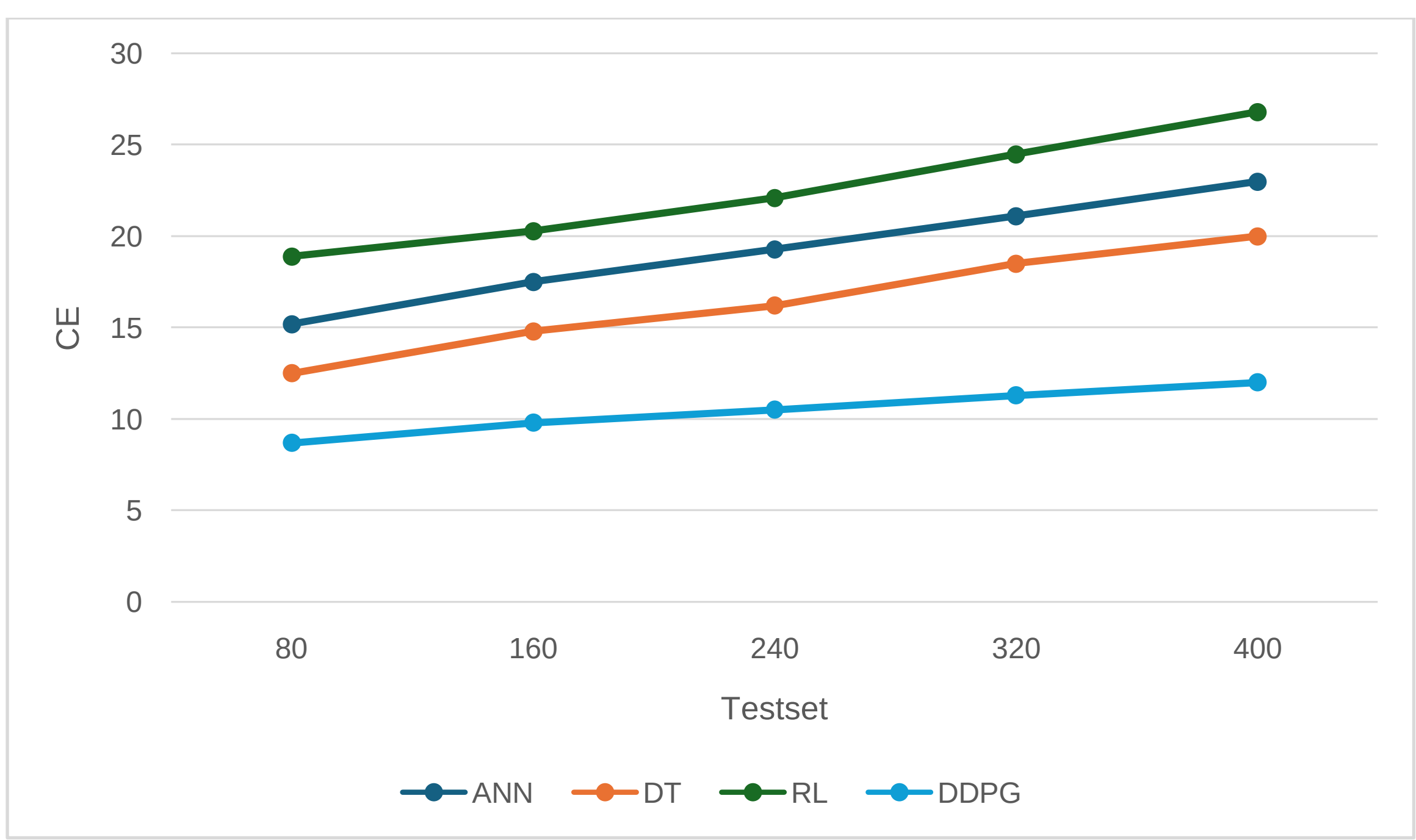

**Figure 6: Computational Efficiency**

The experiments driven by data, as seen in Figures 2–6, reveal how well various strategies work. Across a variety of test data sets, the DDPG method frequently outperforms existing methods (ANN, DT, RL), with an average accuracy improvement of between 10 and 15%. Better precision—an average improvement of 10–15%—is also demonstrated by DDPG over ANN, DT, and RL. This implies that despite decreasing false positives, DDPG more accurately identifies offenders. In a similar spirit, DDPG exhibits ten to 15% higher recall rates than other methods. This suggests that in criminal investigations DDPG catches a larger proportion of actual perpetrators. The F-measure combines recall and precision and shows that DDPG performs 10–15% better overall than existing methods. This implies a reasonable performance in terms of identifying criminals and lowering false positives and a 30–40% improved processing efficiency is achieved by DDPG.

## 5. Conclusion

Experimental results show that DDPG identifying well the offenders in criminal scene investigations when compared with ANN, DT, and RL. DDPG performs well in accuracy, precision, recall, and F-measure and it identify the offenders with reduced false positives precisely with 10-15% improvement. DDPG is effective as it handles faster than ANN, DT, and RL and an 30% improvement in criminal investigations enables quicker decision-making, which increases efficiency.

## References

[1]. Namatēvs, I. (2017). Deep reinforcement learning on HVAC control. *Information Technology and Management Science*, 20(1), 40-45.

[2]. Hanumaiah, V., &Genc, S. (2021). Distributed multi-agent deep reinforcement learning framework for whole-building HVAC control. *arXiv preprint arXiv:2110.13450*.

[3]. Li, W., Zhang, H., van Vlijmen, B., Dechent, P., & Sauer, D. U. (2021). Forecasting battery capacity and power degradation with multi-task learning. *arXiv preprint arXiv:2111.14937*.

[4]. Lin, J., Zhang, Y., &Khoo, E. (2021). Hybrid physics-based and data-driven modeling with calibrated uncertainty for lithium-ion battery degradation diagnosis and prognosis. *arXiv preprint arXiv:2110.13661*.

[5]. Wong, W., Dutta, P., Voicu, O., Chervonyi, Y., Paduraru, C., & Luo, J. (2022). Optimizing industrial HVAC systems with hierarchical reinforcement learning. *arXiv preprint arXiv:2209.08112*.

[6]. Li, W., Zhang, H., van Vlijmen, B., Dechent, P., & Sauer, D. U. (2021). Forecasting battery capacity and power degradation with multi-task learning. *arXiv preprint arXiv:2111.14937*.

[7]. Hanumaiah, V., &Genc, S. (2021). Distributed multi-agent deep reinforcement learning framework for whole-building HVAC control. *arXiv preprint arXiv:2110.13450*.

[8]. Wong, W., Dutta, P., Voicu, O., Chervonyi, Y., Paduraru, C., & Luo, J. (2022). Optimizing industrial HVAC systems with hierarchical reinforcement learning. *arXiv preprint arXiv:2209.08112*.

[9]. Guo, M., Goel, D., Wang, G., Guo, R., Sakurai, Y., & Babar, M. A. (2022). Mechanism design for public projects via three machine learning based approaches. *Autonomous Agents and Multi-Agent Systems*, 36(1), 1-30.

[10]. Xing, Q., Wang, J., Jiang, H., & Wang, K. (2022). Research of a novel combined deterministic and probabilistic forecasting system for air pollutant concentration. *Expert Systems with Applications*, 187, 115919.

[11]. Koh, D., Mishra, A., &Terao, K. (2022). Deep neural network uncertainty quantification for LArTPC reconstruction. *Journal of Instrumentation*, 17(12), P12013.

[12]. Raileanu, G., & de Jong, J. S. (2022). Electrocardiogram interpretation using artificial intelligence: Diagnosis of cardiac and extracardiac pathologic conditions. How far has machine learning reached?.*Current Problems in Cardiology*, 47(6), 100921.